\documentclass[11pt]{article}

\usepackage[final]{acl}

\usepackage{times}
\usepackage{latexsym}
\usepackage{amsmath}
\usepackage[T1]{fontenc}

\usepackage[utf8]{inputenc}

\usepackage{microtype}

\usepackage{inconsolata}

\usepackage{graphicx}

\newcommand{\sd}[1]{\textsubscript{$\pm$#1}}
\title{Do Motion Tokenizers for Co-Speech Gesture Generation \\ Encode Gesture Semantics?
}

\author{
 \textbf{Varsha Suresh\textsuperscript{1,2}\thanks{These authors contributed equally to this work.}}, \textbf{Divij Jain \textsuperscript{3}}\footnotemark[1]\thanks{Work done during an internship at Max Planck Institute for Informatics.}, \textbf{Jia Liu\textsuperscript{1}}, \textbf{M. Hamza Mughal\textsuperscript{2}}, \textbf{Vera Demberg\textsuperscript{1,2}}
\\
\\
 \textsuperscript{1}Saarland University,
 \textsuperscript{2}Max Planck Institute for Informatics, Saarland Informatics Campus,\\
 \textsuperscript{3}IIT Delhi
\\
\small{
vsuresh@mpi-inf.mpg.de, 
cs5240052@iitd.ac.in},\\
\small{jili00008@stud.uni-saarland.de, mmughal@mpi-inf.mpg.de, vera@lst.uni-saarland.de}
}

\usepackage{float}
\begin{document}
\maketitle
\begin{abstract}
Discrete motion tokenizers encode motion as atomic units and are widely used for co-speech gesture generation. It remains unclear which motion properties, especially those relevant to gesture semantics, are recoverable from these codebooks. We probe a reconstruction-trained codebook using 19 co-speech gesture descriptors spanning from raw motion to abstract communicative function. Results show that geometry and handedness are readily decodable from token embeddings, while motion category is only weakly decoded despite showing systematic differences in discrete code usage. This gap between reconstruction quality and descriptor decodability suggests that reconstruction objectives alone do not guarantee that gesture semantics are captured, and that evaluating codebooks on such properties can guide the design of more semantic motion tokenizers.
\end{abstract}

\section{Introduction}

Co-speech gesture generation increasingly relies on discrete motion codebooks, representing continuous motion as sequences of learned tokens \citep{oord2017vqvae, zhang2023t2mgpt, jiang2023motiongpt, liu2024emage}. These codebooks are trained with reconstruction objectives that average error across all joints and frames, favoring motion that is natural and fluid rather than motion that is semantically distinctive. Fine-grained movement, such as finger configuration, contributes only marginally to this loss, since it is overshadowed by larger-magnitude body motion, even though it can be communicatively important \citep{brentari1998prosodic}. Since higher-level properties like motion category and communicative intent are partly expressed through such fine-grained motion, it is unclear whether these atomic units of movement are preserved in the codebook, which is subsequently used for co-speech gesture generation.

Co-speech gesture spans multiple levels of abstraction, from raw motion coordinates to descriptors such as spatial form, handedness, handshape, and communicative function \citep{mcneill1992hand, kendon2004gesture, brentari1998prosodic}. Many of these higher-level descriptors are essential for a gesture to be semantically meaningful \citep{mcneill1992hand, kendon2004gesture}, yet it is unclear whether a codebook trained purely for motion reconstruction preserves them. In this paper, we aim to examine which of these properties are present in a reconstruction-trained gesture codebook, and which are lost.

To study this, we use the semantic gesture dataset from \citet{ram2025gesturecoach}. Building on this dataset and prior gesture annotation frameworks \citep{kipp2007annotation, hensel2025richly}, we define a panel of 19 descriptors spanning raw motion to abstract communicative function. Some descriptors, such as joint-level geometry, are computed automatically from skeleton coordinates, while others, including handshape, handedness, and communicative intent, are manually annotated to capture aspects of gesture that are not directly recoverable from raw coordinates. Following prior work on probing learned representations \citep{alain2016understanding, belinkov2022probing, hewitt2019designing, pimentel2020information}, we use these descriptors to characterize which gesture properties are recoverable from the codebook at two complementary levels: we apply linear and shallow nonlinear probes to mean-pooled token embeddings, and we compare code-activation distributions across descriptor classes to examine whether properties are reflected in patterns of discrete code usage.

Geometry and handedness are strongly decodable from pooled token representations, whereas handshape, motion category, and communicative intent are only weakly decoded. Motion category, however, shows systematic differences in discrete code usage. Handshape remains weakly recoverable under both analyses, suggesting that fine finger configurations, such as a cupped or pinched hand, are largely lost at the tokenization stage. We observe the same pattern on a second, architecturally distinct codebook, EMAGE \citep{liu2024emage}, indicating that these results are not specific to a single tokenizer.

Our contributions are: (1) we introduce a systematic probing framework spanning 19 descriptors, from raw motion to communicative function, to characterize what reconstruction-trained gesture codebooks encode; and (2) Our results suggest that fine-grained, semantically important motion, such as handshape, may not be adequately preserved in a reconstruction-trained gesture codebook, even when the codebook achieves strong overall reconstruction accuracy.

\section{Related Work}
\subsection{Gesture representations and descriptors}
Gesture research has long analyzed movement form alongside communicative function \citep{efron1941gesture,mcneill1992hand,kendon2004gesture,goldinmeadow2003hearing}. Computational approaches utilize raw trajectories \citep{kucherenko2019analyzing} to derive kinematic descriptors \citep{ferstl2020predictability}, or rely on manual annotations of location, orientation, handedness, and handshape \citep{kipp2007annotation,hensel2025richly}. These methodologies parallel sign language phonology, which examines location, handshape, and movement \citep{stokoe1960sign,brentari1998prosodic}. At a higher level, gestures are categorized into semantic types (e.g., iconic, metaphoric) \citep{mcneill1992hand,kendon2004gesture,goldinmeadow2003hearing}, which computational models often treat as distinct targets \citep{kucherenko2022multimodal}. Recent models employ discrete tokenizers via vector quantization for general motion \citep{zhang2023t2mgpt,jiang2023motiongpt} and co-speech gesture \citep{liu2024emage}. They are typically trained purely for reconstruction \citep{oord2017vqvae}. While higher-level semantics are sometimes introduced externally \citep{zhang2024semantic,mughal2025retrieving}, reconstruction performance alone does not clarify which interpretable properties the learned tokens inherently encode.

\subsection{Probing learned representations}

Probing tests whether specific properties are recoverable from learned representations \citep{alain2016understanding, belinkov2019analysis}. We use linear probes \citep{hewitt2019designing, pimentel2020information} complemented by code-activation analysis to evaluate interpretable gesture features \citep{ghaleb2024learning}, differentiating between simple accessibility and discrete structural encoding.

\begin{table*}[t]
\centering
\setlength{\tabcolsep}{5pt}
\begin{tabular}{@{}l r cc cc c@{}}
\hline
& & \multicolumn{2}{c}{\textbf{Probing F1}}
& \multicolumn{2}{c}{\textbf{Code usage}}
& \textbf{Reconstruction} \\
\textbf{Descriptor group}
& \textbf{N}
& \textbf{Linear$\uparrow$}
& \textbf{Non-Linear $\uparrow$}
& \textbf{MI $\uparrow$}
& \textbf{Sig.}
& \textbf{$r$(F1, MPJPE) $\downarrow$} \\
\hline
Motion dynamics
& 2 & 0.49\sd{0.07} & 0.52\sd{0.08} & 0.35 & 1/2 & $-0.26$ \\
Spatial geometry
& 9 & \textbf{0.77}\sd{0.01} & \textbf{0.74}\sd{0.03} & 1.07 & 9/9 & $-0.69$ \\
Handedness
& 1 & 0.73\sd{0.03} & 0.66\sd{0.05} & 0.77 & 1/1 & $\mathbf{-0.89}$ \\
Annotated spatial form
& 4 & 0.30\sd{0.01} & 0.31\sd{0.02} & 1.06 & 3/4 & $-0.48$ \\
Handshape
& 1 & 0.16\sd{0.03} & 0.20\sd{0.07} & 1.50 & 0/1 & $-0.67$ \\
Motion pattern and function
& 2 & 0.11\sd{0.02} & 0.12\sd{0.01} & \textbf{3.62}$^{*}$ & 1/2 & $-0.45$ \\
\hline
\end{tabular}
\caption{Group-level descriptor analysis. Linear and MLP probes use
mean-pooled token embeddings; F1 is reported as mean~$\pm$~SD across
descriptors in each group. MI measures information in discrete code
activations, and Sig.\ reports the number of descriptors with significant
JS segregation. More negative correlations indicate
that lower MPJPE is associated with higher probe performance. $^{*}$This group-level MI is
driven primarily by motion category (MI~=~3.8, $q$~=~0.006)}
\label{tab:results-summary}
\end{table*}


\section{Data and Method}

\subsection{Descriptor Panel}
We analyze the 19 gesture descriptors provided with the
\textsc{Semantix} dataset introduced by
\citet{suresh2026semantic}. We use the subset of 778 semantic gesture
clips from TED, which originates from the gesture regions identified by
\citet{ram2025gesturecoach}. Each clip is associated with descriptors
covering physical gesture form and communicative function. These include
automatically computed descriptors derived from the 3D skeleton motion,
as well as manually annotated descriptors such
as handshape, motion category class, and communicative function. In this
paper, we retain the descriptor definitions and labels from the original
study and organize the 19 descriptors into six groups to support our
group-level probing and code-usage analyses. Table~\ref{tab:descriptor-panel} summarizes the panel; full per-descriptor class labels are provided in Appendix~\ref{sec:appendix}.

\begin{table}[h]
\centering
\resizebox{\linewidth}{!}{
\begin{tabular}{ll}
\hline
\textbf{Group} & \textbf{Descriptor} \tabularnewline
\hline
Motion dynamics & Gesture size, Kinematics \tabularnewline
Spatial geometry & Left/Right vertical, horizontal, depth, \tabularnewline
& Left/Right elbow bend, Hand reach \tabularnewline
Handedness & Handedness \tabularnewline
Annotated spatial form & Hand orientation, Spatial radial, \tabularnewline
& Spatial height, Spatial distance \tabularnewline
Handshape & Handshape \tabularnewline
Motion pattern and function & Motion category, Comm. intent \tabularnewline
\hline
\end{tabular}}
\caption{Descriptor panel used in the main analysis.}
\label{tab:descriptor-panel}
\end{table}

\subsection{Codebook Architecture and Training}\label{sec:models}
We analyze a two-stream vector-quantized auto-encoder (VQ-VAE) \citep{oord2017vqvae} over a 38-joint 3D upper-body skeleton, partitioned into \emph{body} and \emph{hand} streams. Models are trained purely to minimize a position reconstruction and VQ commitment loss on a combined BEAT and TED gesture corpus. Our primary probing analyzes the best-reconstructing configuration: latent dimension $L=256$, codebook size $K=2048$, and temporal downsampling $d=4$. Full architectural, configuration sweep, and training details are provided in Appendix~\ref{sec:appendix-architecture}.

\subsection{Probing Setup}

We probe the codebook at two complementary levels: sequence-level probing of pooled token representations and code-level analysis of discrete code-usage patterns.

\paragraph{Linear and nonlinear probing.}
Each gesture is represented using either mean-pooled codebook embeddings or normalized one-hot code histograms. For each descriptor, we train multinomial logistic regression with balanced class weights and report 5-fold cross-validated macro-F1 against majority-class and random-assignment baselines. We additionally train a shallow MLP with one 128-unit ReLU layer and $L_2$ regularization on the pooled embeddings, using identical folds and class balancing. This tests whether weak linear performance reflects nonlinear encoding while limiting probe capacity \citep{hewitt2019designing,pimentel2020information}.

\paragraph{Code-level activation analysis.}
Each clip is represented as a normalized code-frequency histogram. We measure class separation using the mean pairwise Jensen--Shannon divergence between class-average usage profiles \citep{lin1991divergence}. Significance is assessed using label-permutation tests with Benjamini--Hochberg correction across descriptors within each stream. As a complementary measure, we sum the mutual information between each active code's presence and the descriptor label \citep{yang1997comparative}. Further details are provided in Appendix~\ref{sec:appendix-activation-proc}.

\paragraph{Reconstruction fidelity.}
We test whether descriptor accessibility tracks reconstruction quality by correlating macro-F1 with mean per-joint position error (MPJPE) across models varying latent dimension, codebook size, and temporal downsampling. Further details are provided in Appendix~\ref{sec:appendix-architecture}.

\section{Results}
\label{sec:results}

\subsection{Gesture Descriptor Probing}
\label{sec:descriptor-probing}

Table~\ref{tab:results-summary} summarizes descriptor accessibility from
pooled token representations, structure in discrete code usage, and its
relationship with reconstruction fidelity.

\begin{figure}[h]
\centering
\includegraphics[width=\columnwidth]{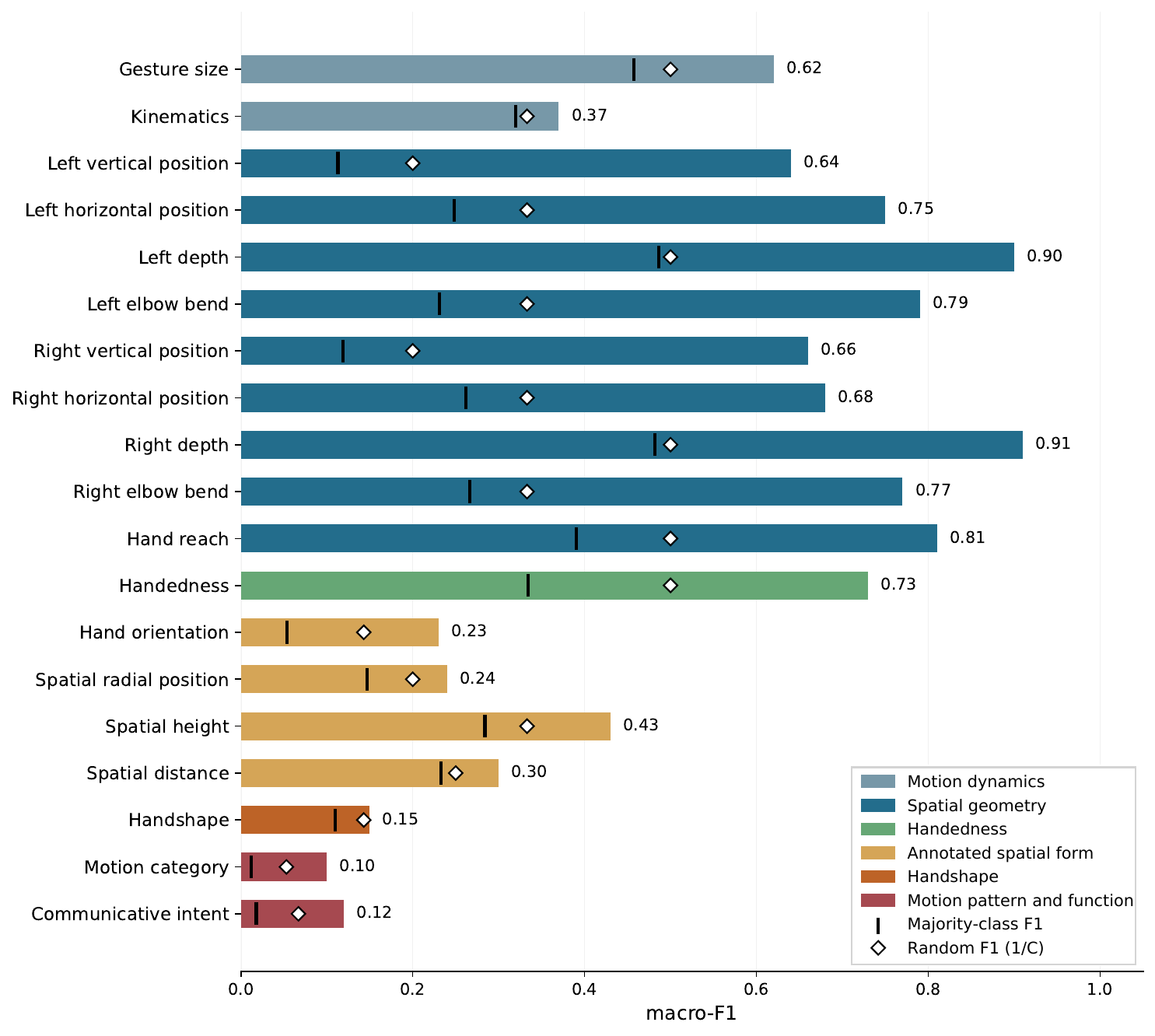}
\caption{Probe macro-F1 for each descriptor}
\label{fig:baseline-probe}
\end{figure}

\paragraph{Accessibility from pooled representations.}
Probe performance shows a clear gradient from motion-grounded properties
to higher-level gesture descriptors. Spatial geometry and handedness
are the most accessible groups, reaching macro-F1 scores of $0.768$ and
$0.734$, respectively. Performance decreases for annotated spatial form ($0.299$),
hand configuration ($0.155$), and motion pattern and function ($0.113$). The descriptor-level comparison in
Figure~\ref{fig:baseline-probe} shows the same pattern: geometric
properties such as left and right depth exceed $0.89$ macro-F1, whereas
handshape, motion category, and communicative intent remain close to
their chance baselines. A shallow MLP does not change this ordering;
motion pattern and function remains effectively unchanged
($0.113\rightarrow0.112$). Its weak linear accessibility is therefore
unlikely to reflect shallow nonlinear encoding alone.

\paragraph{Structure in discrete code usage.}
Activation analysis asks whether descriptor classes use different subsets
of codes. Higher MI indicates that code activations are more informative
about the descriptor, while $q<0.05$ indicates reliable separation after
multiple-testing correction. Spatial geometry is significant for all
nine descriptors, and handedness is also reliably separated. Most
notably, motion pattern and function has the highest group-level MI ($3.62$),
driven by motion category (MI $=3.8$, $q=0.006$), despite its weak
pooled probing performance. This suggests that motion category is encoded
in \emph{which} codes are used, but is obscured by pooling. This does not hold for all higher-level properties: communicative intent
($q=0.06$) and handshape ($q=0.70$) show no reliable overall
separation. Thus, weak pooled probing reflects hidden code-level structure
for motion category, but limited evidence for handshape and most
communicative intents.

\paragraph{Relationship with reconstruction metrics}
Negative correlations indicate that lower reconstruction error is
associated with higher probe performance. This relationship is strongest
for handedness ($r=-0.88$) and spatial geometry ($r=-0.69$),
suggesting that reconstruction quality mainly tracks motion-grounded
properties.

\subsection{Generalization to Another Codebook}
\label{sec:emage}

\begin{figure}[H]

\centering
\includegraphics[width=1.0\linewidth]{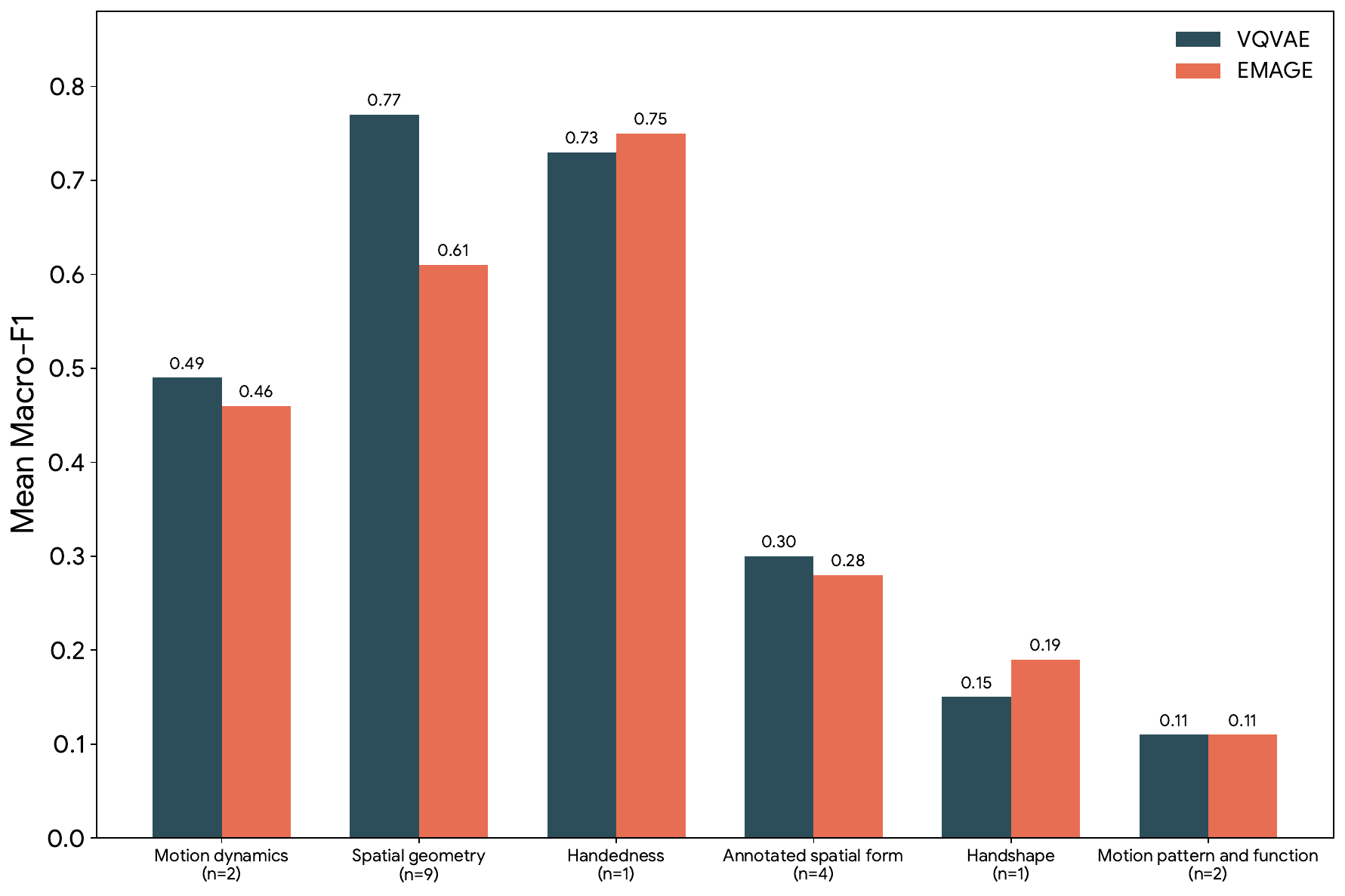}
\caption{Probing macro-F1 by descriptor group, for EMAGE.}
\label{fig:emage-comparison}
\end{figure}

We repeat the analysis on EMAGE \citep{liu2024emage}.
As shown in Figure~\ref{fig:emage-comparison}, both
codebooks exhibit the same accessibility gradient:
spatial geometry is strongest, while properties such
as handshape remain weak. Although absolute scores
differ slightly, the ordering of descriptor groups is
stable across architectures. This consistency suggests
that the accessibility gradient is not specific to a
single tokenizer design.

\begin{figure}[h]
    \centering
\includegraphics[width=\linewidth]{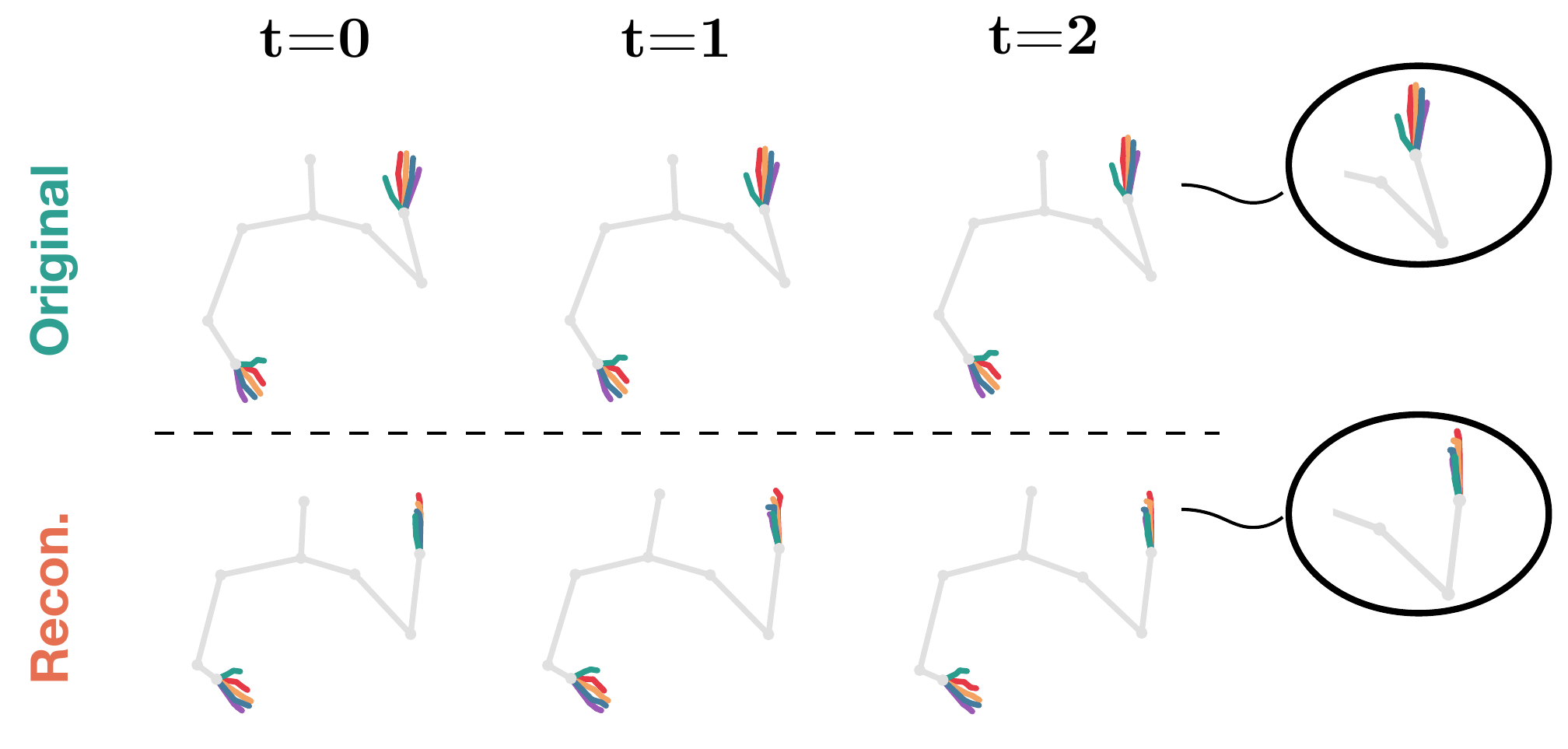}
    \caption{Qualitative comparison showing that reconstruction-only VQ-VAEs preserve upper-body geometry but average out fine-grained hand poses}
\label{fig:case-study}
\end{figure}

\section{Qualitative Case Study}
Figure~\ref{fig:case-study} illustrates a pattern that corroborates probing performance on communicative functions. VQ-VAE based tokenizers trained solely with a reconstruction loss, learn representations that decode into averaged-out motions. In particular, hands do not achieve same fidelity at reconstruction as compared to upper body joints, due to difference of scale between hands and upper body. Therefore, better performance on upper body results in strong accessibility of geometric descriptors. In contrast, poorly reconstructed hand poses depict different communicative functions, which eventually leads to subpar performance on gesture functions. These findings highlight that reconstruction losses do not incentivize learning of gesture semantics as part of the representation learning in codebooks. 

\section{Conclusion}
\label{sec:conclusion}

Across two codebook architectures, geometric properties and handedness are readily recoverable, while handshape, motion category, and communicative intent remain weak. Better reconstruction and finer temporal resolution mainly improve motion-grounded properties, suggesting that reconstruction quality alone is not a sufficient measure of semantic accessibility. Gesture codebooks should therefore be evaluated using both probing and code-level analyses, and future tokenizers should explicitly preserve communicatively relevant properties.

\section*{Limitations}

Our analysis is limited to 778 semantic gestures from
TED and two skeleton-based, reconstruction-trained
tokenizers, so the observed accessibility gradient may
not generalize to other domains, motion representations,
or tokenizer objectives. In addition, we did not ablate the composition of the training corpus as the codebook was trained on BEAT and TED jointly, so the contribution of each dataset to descriptor accessibility remains untested. Manual descriptors, particularly
handshape and communicative intent, are relatively sparse
and may contain annotation subjectivity. Moreover, low
probe performance indicates limited recoverability under
the selected pooled representations and probe families;
it does not establish that a property is entirely absent
or inaccessible to a sequence-aware model. Future work
should evaluate larger and more diverse datasets, use
temporal probes, and test whether descriptor accessibility
predicts the semantic quality of generated gestures.

\section*{Acknowledgments}
This work is funded in part by the Deutsche Forschungsgemeinschaft (DFG, German Research Foundation) - GRK 2853/1 “Neuroexplicit Models of Language, Vision, and Action” - project number 471607914."

\bibliography{custom}

\appendix
\section{Appendix}
\label{sec:appendix}
\subsection{Additional Details: Data and Method}
\subsubsection{Descriptor Classes}
Table~\ref{tab:descriptor-panel-appendix} shows the detailed classes in each descriptor.
\begin{table*}[t]
\centering
\scriptsize
\setlength{\tabcolsep}{3pt}
\renewcommand{\arraystretch}{1.08}
\resizebox{\linewidth}{!}{%
\begin{tabular}{llccl}
\hline
\textbf{Group}
& \textbf{Descriptor}
& \textbf{Annotation}
& \textbf{$C$}
& \textbf{Labels} \\
\hline

Motion dynamics
& Gesture size
& Manual
& 2
& normal; big \\

Motion dynamics
& Kinematics
& Manual
& 3
& slow; normal; fast \\

Spatial geometry
& Left vertical position
& Automatic
& 5
& below-waist; waist-level; torso-level; chest-level; above-head \\

Spatial geometry
& Left horizontal position
& Automatic
& 3
& body-center; torso-side; extended-outward \\

Spatial geometry
& Left depth
& Automatic
& 2
& at-torso; in-front-of-torso \\

Spatial geometry
& Left elbow bend
& Automatic
& 3
& straight; slightly-bent; bent \\

Spatial geometry
& Right vertical position
& Automatic
& 5
& below-waist; waist-level; torso-level; chest-level; above-head \\

Spatial geometry
& Right horizontal position
& Automatic
& 3
& body-center; torso-side; extended-outward \\

Spatial geometry
& Right depth
& Automatic
& 2
& at-torso; in-front-of-torso \\

Spatial geometry
& Right elbow bend
& Automatic
& 3
& straight; slightly-bent; bent \\

Spatial geometry
& Hand reach
& Automatic
& 2
& mid-reach; extended \\

Handedness
& Handedness
& Manual
& 2
& one hand; both hands \\

Annotated spatial form
& Hand orientation
& Manual
& 7
& inward; outward; toward center; up; down; side; forward \\

Annotated spatial form
& Spatial radial position
& Manual
& 5
& inward; front; side; out; far out \\

Annotated spatial form
& Spatial height
& Manual
& 3
& belt; chest; head \\

Annotated spatial form
& Spatial distance
& Manual
& 4
& touch; close; normal; far \\

Handshape
& Handshape
& Manual
& 7
& palm; cup; pinch; fist; bunch; fingers; index finger \\ \\ 

Motion pattern and function
& Motion category
& Manual
& 19
& \shortstack[l]{
arc-like trajectory; counting; diagonal; flick; glide; horizontal;\\
long-tail; pointing; pull; push; rotate; shake; step; strike; sway;
sweep; vertical down; vertical up; wave
} \\ \\ 

Motion pattern and function
& Communicative intent
& Manual
& 15
& \shortstack[l]{
contrast; discourse; emotion; emphasis; enumeration; listing;
negation; other-reference;\\
progression; quantification; relativity; self-reference; symbolise;
temporal reference; uncertainty
} \\

\hline
\end{tabular}
}
\caption{The 19-descriptor panel used for probing. Automatic descriptors
are computed and discretized from skeleton coordinates, whereas manual
descriptors are human annotated. $C$ denotes the number of retained
classes after removing labels represented by fewer than five clips.}
\label{tab:descriptor-panel-appendix}
\end{table*}

\subsubsection{Codebook Architecture and Training}
\label{sec:appendix-architecture}
We analyse a two-stream vector-quantised auto-encoder over a 38-joint 3D upper-body skeleton. The joints are partitioned into a \emph{body} stream of 8 coarse joints and a \emph{hand} stream of 30 finger joints. The two streams have separate encoders and codebooks and are quantised independently; a single shared decoder reconstructs the full pose. The hand joints are expressed relative to their wrist before encoding, meaning the hand stream encodes wrist-local finger configuration while the body stream encodes coarse-joint positions.

Each stream's encoder is a 1D temporal convolutional stack (kernel 4, stride 2) that downsamples time by a factor $d$. Each codebook is a vector-quantised bank trained with exponential-moving-average updates (decay $0.99$) and dead-code reinitialisation. Tokenisation processes each clip in non-overlapping $d$-frame chunks, yielding one token per stream per chunk.

To map how representational granularity affects descriptor accessibility, we swept a $3\times3\times3$ grid of 27 models over $L\in\{64,128,256\}$, $K\in\{1024,2048,4096\}$, and $d\in\{4,8,16\}$. Every model is trained purely to reconstruct motion. The codebook is trained on a combined 3D gesture-motion corpus comprising the BEAT and TED datasets (processed at native frame rates of 30 FPS and 24--30 FPS respectively, balanced by duration). Models are optimised with the Adam optimiser (learning rate $1 \times 10^{-3}$, batch size 8) for up to 25 epochs (with early stopping), minimising a position reconstruction loss plus the VQ commitment loss weighted by $\beta=1.0$.

\subsubsection{Probing Setup}\label{sec:appendix-baselines}
Table~\ref{tab:linear-baselines-appendix} reports descriptor-level linear probing results. The main pattern is consistent with the group-level results in the main paper. Automatic geometric descriptors and handedness are the strongest, while handshape, motion category, and communicative intent remain weak.

A notable trend across these descriptors is that the continuous codebook vector (Embedding) outperforms the code histogram (One-hot) for almost all descriptors. The advantage is largest on graded spatial/positional descriptors (e.g., vertical reach bias, spatial distance), where a continuous vector successfully captures gradation that a discrete histogram naturally discards. One-hot representations only edge ahead on a handful of low-signal descriptors (like handshape), where the absolute F1 is already near the noise floor.

\subsection{Additional Results}
\subsubsection{Code-Level Activation Procedure \& Results}
\label{sec:appendix-activation-proc}
To rigorously evaluate whether a descriptor is reflected in the use of individual codebook entries, we employ an information-theoretic approach.

Jensen-Shannon divergence provides a symmetric and bounded metric for scoring how much the classes' profiles differ. The summed presence Mutual Information (MI) provides an additive measure of how much information the active vocabulary collectively carries about the label, avoiding the combinatorial infeasibility of estimating a full joint contingency table $I(Y;K)$ over all 2048 codes.

Table~\ref{tab:activation-descriptor-appendix} reports the descriptor-level results. The data reveals that weak linear performance does not possess a single interpretation. Some descriptors leave strong code-level traces despite failing linear probing (e.g., motion category, $q=0.006$), indicating the information lives in \emph{which} discrete units are activated rather than their averaged vector geometry. Conversely, other abstract properties like communicative intent ($q=0.063$) and handshape ($q=0.697$) fail to reach broad code-level significance, indicating that these features are genuinely absent from the token representation.

\subsubsection{Further examples for the code-level metrics.}
To clarify the code-level metrics, we describe what they return in two extreme cases and in two real descriptors. In the first extreme, both classes of a binary descriptor use every code equally often. The two class-average profiles are then identical, so $\text{JS} = 0$, and because knowing which code fired tells us nothing about the class, $\text{MI} = 0$ as well. In the opposite extreme, each class uses only its own code and never the other's. The profiles share no codes, so JS reaches its maximum of $1$ bit, and observing a single code reveals the class with certainty, so MI is maximal. 

Real descriptors sit between these two extremes. For handedness, both classes share generic resting codes, but ``one hand'' clips concentrate on a few specific codes (e.g., hand code 746) while ``both hands'' clips use a separate set of bimanual codes. This partial overlap gives intermediate values ($\text{JS} = 0.31$, $\text{MI} = 0.766$), and the permutation test confirms the separation is far above what random labels would produce ($q = 0.006$). Left depth classes are heavily imbalanced (94\% vs.\ 6\%), so the minority profile is estimated from very few clips and looks spiky. JS weighs both profiles equally regardless of how many clips they come from, so it becomes inflated ($\text{JS} = 0.83$). MI instead accounts for how rare the minority class is, so it stays moderate ($\text{MI} = 0.493$). Reading JS together with its permutation $q$-value and MI therefore prevents false discoveries caused by class imbalance.

\begin{table*}[t]
\centering
\small
\begin{tabular}{p{0.20\textwidth}|p{0.26\textwidth}|r|r|r|r|r}
\hline
\textbf{Group} & \textbf{Descriptor} & \textbf{Cls.} &
\textbf{Maj-F1} & \textbf{Rand-F1} & \textbf{Embed.} &
\textbf{One-hot} \tabularnewline
\hline

Motion dynamics & Gesture size & 2 & 0.457 & 0.500 &
\textbf{0.619} & 0.562 \tabularnewline

Motion dynamics & Kinematics & 3 & 0.320 & 0.333 &
\textbf{0.367} & 0.340 \tabularnewline

\hline

Spatial geometry & Left vertical position & 5 & 0.113 & 0.200 &
\textbf{0.639} & 0.563 \tabularnewline

Spatial geometry & Left horizontal position & 3 & 0.248 & 0.333 &
\textbf{0.749} & 0.611 \tabularnewline

Spatial geometry & Left depth & 2 & 0.486 & 0.500 &
\textbf{0.898} & 0.884 \tabularnewline

Spatial geometry & Left elbow bend & 3 & 0.231 & 0.333 &
\textbf{0.787} & 0.669 \tabularnewline

Spatial geometry & Right vertical position & 5 & 0.119 & 0.200 &
\textbf{0.663} & 0.467 \tabularnewline

Spatial geometry & Right horizontal position & 3 & 0.262 & 0.333 &
\textbf{0.679} & 0.566 \tabularnewline

Spatial geometry & Right depth & 2 & 0.482 & 0.500 &
\textbf{0.912} & 0.858 \tabularnewline

Spatial geometry & Right elbow bend & 3 & 0.266 & 0.333 &
\textbf{0.772} & 0.487 \tabularnewline

Spatial geometry & Hand reach & 2 & 0.390 & 0.500 &
\textbf{0.809} & 0.751 \tabularnewline

\hline

Handedness & Handedness & 2 & 0.334 & 0.500 &
\textbf{0.734} & 0.664 \tabularnewline

\hline

Annotated spatial form & Hand orientation & 7 & 0.054 & 0.143 &
\textbf{0.227} & 0.201 \tabularnewline

Annotated spatial form & Spatial radial position & 5 & 0.147 & 0.200 &
\textbf{0.236} & 0.228 \tabularnewline

Annotated spatial form & Spatial height & 3 & 0.284 & 0.333 &
0.434 & \textbf{0.438} \tabularnewline

Annotated spatial form & Spatial distance & 4 & 0.233 & 0.250 &
\textbf{0.300} & 0.297 \tabularnewline

\hline

Handshape & Handshape & 7 & 0.110 & 0.143 &
0.155 & \textbf{0.197} \tabularnewline

\hline

Motion pattern and function & Motion category & 19 & 0.012 & 0.053 &
\textbf{0.105} & 0.083 \tabularnewline

Motion pattern and function & Communicative intent & 15 & 0.018 & 0.067 &
\textbf{0.121} & 0.101 \tabularnewline

\hline
\end{tabular}

\caption{Descriptor-level linear probing macro-F1 with chance baselines.
\textbf{Cls.} is the number of retained classes ($\geq 5$ samples).
\textbf{Maj-F1} is the majority-class macro-F1 and \textbf{Rand-F1} is
the random-assignment macro-F1 ($=1/C$ in expectation). For
near-balanced binary labels such as depth and handedness, Rand-F1
exceeds Maj-F1, so the higher of the two is the relevant chance level.
The best of the two probe representations is shown in bold. Spatial
geometry and handedness clear chance by large margins, while motion
category, communicative intent, and handshape exceed chance only
marginally.}

\label{tab:linear-baselines-appendix}
\end{table*}

\begin{table*}[t]
\centering
\small
\begin{tabular}{p{0.24\textwidth}|p{0.30\textwidth}|r|l|r|r}
\hline
\textbf{Group} & \textbf{Descriptor} & \textbf{Cls.} &
\textbf{Stream} & \textbf{MI} & \textbf{$q$} \tabularnewline
\hline

Motion dynamics & Gesture size & 2 & hand & 0.397 &
\textbf{0.006} \tabularnewline

Motion dynamics & Kinematics & 3 & hand & 0.311 &
0.327 \tabularnewline

\hline

Spatial geometry & Left vertical position & 5 & hand & 1.849 &
\textbf{0.006} \tabularnewline

Spatial geometry & Left horizontal position & 3 & hand & 1.056 &
\textbf{0.006} \tabularnewline

Spatial geometry & Left depth & 2 & hand & 0.493 &
\textbf{0.006} \tabularnewline

Spatial geometry & Left elbow bend & 3 & hand & 1.039 &
\textbf{0.006} \tabularnewline

Spatial geometry & Right vertical position & 5 & hand & 1.851 &
\textbf{0.006} \tabularnewline

Spatial geometry & Right horizontal position & 3 & hand & 1.024 &
\textbf{0.006} \tabularnewline

Spatial geometry & Right depth & 2 & hand & 0.575 &
\textbf{0.006} \tabularnewline

Spatial geometry & Right elbow bend & 3 & hand & 0.875 &
\textbf{0.006} \tabularnewline

Spatial geometry & Hand reach & 2 & hand & 0.865 &
\textbf{0.006} \tabularnewline

\hline

Handedness & Handedness & 2 & hand & 0.766 &
\textbf{0.006} \tabularnewline

\hline

Annotated spatial form & Hand orientation & 7 & hand & 1.802 &
\textbf{0.006} \tabularnewline

Annotated spatial form & Spatial radial position & 5 & hand & 1.191 &
\textbf{0.006} \tabularnewline

Annotated spatial form & Spatial height & 3 & hand & 0.534 &
0.208 \tabularnewline

Annotated spatial form & Spatial distance & 4 & hand & 0.697 &
\textbf{0.021} \tabularnewline

\hline

Handshape & Handshape & 7 & hand & 1.499 &
0.697 \tabularnewline

\hline

Motion pattern and function & Motion category & 19 & hand & 3.790 &
\textbf{0.006} \tabularnewline

Motion pattern and function & Communicative intent & 15 & hand & 3.450 &
0.063 \tabularnewline

\hline
\end{tabular}

\caption{Descriptor-level code activation results using the stream with
the highest mutual information for each descriptor. Bold $q$-values
indicate FDR-corrected $q<0.05$. Spatial geometry is consistently
significant. Motion category is the main higher-level descriptor that
shows a reliable code-level trace despite weak linear probing
performance.}

\label{tab:activation-descriptor-appendix}
\end{table*}


\end{document}